\documentclass[runningheads]{llncs}
\usepackage{graphicx}
\usepackage{IEEEtrantools}
\usepackage{amssymb}
\usepackage{amsmath}
\usepackage{subfigure}
\usepackage{algorithm}
\usepackage{algorithmic}

\newcommand{\dd}{\mathop{}\!\mathrm{d}}

\newtheorem{result}{Result}

\begin{document}
\title{Membership-Mappings for Data Representation Learning: Measure Theoretic Conceptualization\thanks{Supported by the Austrian Research Promotion Agency (FFG) Sub-Project PETAI (Privacy Secured Explainable and Transferable AI for Healthcare Systems); the Federal Ministry for Climate Action, Environment, Energy, Mobility, Innovation and Technology (BMK); the Federal Ministry for Digital and Economic Affairs (BMDW); and the Province of Upper Austria in the frame of the COMET - Competence Centers for Excellent Technologies Programme managed by Austrian Research Promotion Agency FFG.}}
\titlerunning{Measure Theoretic Conceptualization of Membership-Mappings}
%
\author{Mohit Kumar\inst{1,2} \and
Bernhard Moser\inst{1} \and
Lukas Fischer\inst{1} \and
Bernhard Freudenthaler\inst{1}}
\authorrunning{M. Kumar et al.}
%
\institute{Software Competence Center Hagenberg GmbH, A‑4232 Hagenberg	
Austria 
\email{mohit.kumar@scch.at}  \and
Institute of Automation, Faculty of Computer Science and Electrical Engineering, 
University of Rostock, Germany}
\maketitle              
\begin{abstract}
A fuzzy theoretic analytical approach was recently introduced that leads to efficient and robust models while addressing automatically the typical issues associated to parametric deep models. However, a formal conceptualization of the fuzzy theoretic analytical deep models is still not available. This paper introduces using measure theoretic basis the notion of \emph{membership-mapping} for representing data points through attribute values (motivated by fuzzy theory). A property of the membership-mapping, that can be exploited for data representation learning, is of providing an interpolation on the given data points in the data space. An analytical approach to the variational learning of a membership-mappings based data representation model is considered. 
\end{abstract}

\section{Introduction}
Deep neural networks have been successfully applied in a wide range of problems but their training requires a large amount of data. The issues concerning neural networks based parametric deep models include determining the optimal model structure, requirement of large training dataset, and iterative time-consuming nature of numerical learning algorithms. These issues have motivated the development of a nonparametric deep model~\cite{8888203} that is learned analytically for representing data points. The study in~\cite{8888203} introduces the concept of \emph{fuzzy-mapping} which is about representing mappings through a fuzzy set with a membership function such that the dimension of membership function increases with an increasing data size. The main result of~\cite{8888203} is that a deep model formed via a composition of finite number of nonparametric fuzzy-mappings can be learned analytically and the analytical approach leads to a robust and computationally fast method of data representation learning. A core issue in machine learning is rigorously accounting for the uncertainties. While probability theory is widely used to study uncertainties in machine learning, the applications of fuzzy theory in machine learning remain relatively unexplored. Both probability and fuzzy theory have been combined to design stochastic fuzzy systems~\cite{5447695,5759770,7100899}. For an analytical design and analysis of machine learning models, a pure fuzzy theoretic approach was introduced~\cite{KUMAR2017668} where fuzzy membership functions quantifying uncertainties are determined via variational optimization~\cite{Zhang2017}. Although the fuzzy based analytical learning approach to the learning of deep models (as suggested in~\cite{8888203,9216097,KUMAR20211}) leads to the development of efficient and robust machine learning models, a formal conceptualization of the fuzzy theoretic analytical deep models is still not available. Thus, our aim here is to present a measure theoretic conceptualization of fuzzy based analytical deep models.  

The study introduces using measure theoretic basis the concept of \emph{membership-mappings}. The membership-mapping in this study has been referred to a measure theoretic conceptualization of the fuzzy-mapping (previously studied in~\cite{8888203,9216097,KUMAR20211}). The membership-mappings allow a representation of data points through attribute values. This representation is motivated by fuzzy theory where the attributes are linguistic variables. A membership-mapping is characterized by a membership function that evaluates the degree-of-matching of data points to the attribute induced by a sequence of observations. The membership functions have been constrained to be satisfying the properties of a) nowhere vanishing, b) positive and bounded integrals, and c) consistency of induced probability measure. For a set of measurable functions, the membership function induces a probability measure (that is guaranteed by Kolmogorov extension theorem). The expectations w.r.t. the defined probability measure can be calculated via simply computing a weighted average with membership function as the weighting function. Finally, an analytical approach to the variational learning of a membership-mappings based data representation model is considered following~\cite{8888203,9216097,KUMAR20211}.    

\section{Notations and Definitions}
Let $n,N,p,M \in \mathbb{N}$. Let $\mathcal{B}(\mathbb{R}^N)$ denote the \emph{Borel $\sigma-$algebra} on $\mathbb{R}^N$, and let $\lambda^N$ denote the \emph{Lebesgue measure} on $\mathcal{B}(\mathbb{R}^N)$. Let $(\mathcal{X}, \mathcal{A} , \rho)$ be a probability space with unknown probability measure $\rho$. Let $\mathcal{S}$ be the set of finite samples of data points drawn i.i.d. from $\rho$, i.e.,
\begin{IEEEeqnarray}{rCl}  
\mathcal{S} & := &  \{ (x^i  \sim \rho )_{i=1}^N \; | \; N \in \mathbb{N} \}.
\end{IEEEeqnarray} 
For a sequence $\mathrm{x} = (x^1,\cdots,x^N) \in \mathcal{S}$, let $|\mathrm{x}|$ denote the cardinality i.e. $|\mathrm{x}| = N$. If $\mathrm{x} = (x^1, \cdots, x^N),\; \mathrm{a}= (a^1, \cdots, a^M) \in \mathcal{S}$, then $\mathrm{x} \wedge \mathrm{a}$ denotes the concatenation of the sequences $\mathrm{x}$ and $\mathrm{a}$, i.e., $\mathrm{x} \wedge \mathrm{a} = (x^1, \ldots, x^N, a^1, \ldots, a^M)$. $\mathbb{F}(\mathcal{X})$ denotes the set of $\mathcal{A}$-$\mathcal{B}(\mathbb{R})$ measurable functions $f:\mathcal{X} \rightarrow \mathbb{R}$, i.e.,
\begin{IEEEeqnarray}{rCl}  
\mathbb{F}(\mathcal{X}) & := &  \{ f:\mathcal{X} \rightarrow \mathbb{R}  \; | \;  \mbox{$f$ is $\mathcal{A}$-$\mathcal{B}(\mathbb{R})$ measurable}\}.
\end{IEEEeqnarray} 
For convenience, the values of a function $f \in \mathbb{F}(\mathcal{X})$ at points in the collection $\mathrm{x} = (x^1,\cdots,x^N)$ are represented as $f(\mathrm{x})=(f(x^1),\cdots,f(x^N))$. For a given $\mathrm{x} \in \mathcal{S}$ and $A\in \mathcal{B}(\mathbb{R}^{|\mathrm{x}|})$, the cylinder set $\mathcal{T}_{\mathrm{x}}(A)$ in $\mathbb{F}(\mathcal{X})$ is defined as 
\begin{IEEEeqnarray}{rCl}
\mathcal{T}_{\mathrm{x}}(A) &: = & \{ f \in \mathbb{F}(\mathcal{X}) \; | \; f(\mathrm{x}) \in A   \}.
\end{IEEEeqnarray}
Let $\mathcal{T}$ be the family of cylinder sets defined as
\begin{IEEEeqnarray}{rCl}
\mathcal{T} & := & \left\{ \mathcal{T}_{\mathrm{x}}(A)\; | \; A \in \mathcal{B}(\mathbb{R}^{|\mathrm{x}|}),\; \mathrm{x} \in \mathcal{S} \right \}.
\end{IEEEeqnarray} 
Let $\sigma(\mathcal{T})$ be the $\sigma$-algebra generated by $\mathcal{T}$. Given two $\mathcal{B}(\mathbb{R}^N)-\mathcal{B}(\mathbb{R})$ measurable mappings, $g:\mathbb{R}^N \rightarrow \mathbb{R}$ and $\mu:\mathbb{R}^N \rightarrow \mathbb{R}$, the weighted average of $g(\mathrm{y})$ over all $\mathrm{y} \in \mathbb{R}^{N}$, with $\mu(\mathrm{y})$ as the weighting function, is computed as   
\begin{IEEEeqnarray}{rCl}
\label{eq_738118.427179} \left< g \right>_{\mu}& := & \frac{1}{ \int_{\mathbb{R}^{N}}  \mu(\mathrm{y})\, \dd\lambda^{N}(\mathrm{y})} \int_{\mathbb{R}^{N}} g(\mathrm{y}) \mu(\mathrm{y})\, \dd \lambda^{N}(\mathrm{y}).
\end{IEEEeqnarray} 
\section{Representation of Samples via Attribute Values} 
Let us consider a given observation $x \in \mathcal{X}$, a data point $\tilde{x} \in \mathcal{X}$, and a mapping $\mathbf{A}_x(\tilde{x}): \tilde{x} \mapsto \mathbf{A}_x(\tilde{x}) \in [0,1]$ such that $\mathbf{A}_x(\tilde{x})$ can be interpreted as evaluation of the degree to which the data point $\tilde{x}$ matches a given attribute induced by the observation $x$. $\mathbf{A}_x(\cdot)$ is called a membership function and this interpretation is motivated by fuzzy theory. In our approach we consider $\mathbf{A}_{x,f}(\tilde x) = (\zeta_{x} \circ f) (\tilde x)$ to be composed of two mappings $f: \mathcal{X} \rightarrow \mathbb{R}$ and $\zeta_x:  \mathbb{R} \rightarrow [0,1]$. $f \in \mathbb{F}(\mathcal{X})$ can be interpreted as physical measurement (e.g., temperature), and $\zeta_x (f(\tilde x))$ as degree to which $\tilde x$ matches the attribute under consideration, e.g. ``\emph{hot}'' where e.g. $x$ is a representative sample of ``\emph{hot}''. Next, we extend this concept to sequences of data points in order to evaluate how much a sequence 
$\tilde{\mathrm{x}} = (\tilde x^1, \ldots, \tilde x^N) \in \mathcal{S}$ matches to the attribute induced by observed sequence $\mathrm{x} = (x^1, \ldots, x^N) \in \mathcal{S}$ w.r.t. the feature $f$ via defining
\begin{IEEEeqnarray}{rCl}
\mathbf{A}_{\mathrm{x},f}(\tilde{\mathrm{x}}) & = & (\zeta_{\mathrm{x}} \circ f)(\tilde{\mathrm{x}}) \\
& = & \zeta_{\mathrm{x}} (f(\tilde x^1), \ldots, f(\tilde x^N)),
\end{IEEEeqnarray} 
where the membership functions $\zeta_{\mathrm{x}}:\mathbb{R}^{|\mathrm{x}|} \rightarrow [0,1]$, $\mathrm{x} \in \mathcal{S}$, satisfy the following properties:
\begin{description}
    \item[Nowhere Vanishing:] $\zeta_{\mathrm{x}}(\mathrm{y}) > 0$ for all $\mathrm{y} \in \mathbb{R}^{|\mathrm{x}|}$, i.e.,
\begin{IEEEeqnarray}{rCl}
    \label{eq:supp}
     \mbox{supp}[\zeta_{\mathrm{x}}] & = & \mathbb{R}^{|\mathrm{x}|}.
\end{IEEEeqnarray}   
    \item[Positive and Bounded Integrals:] the functions $\zeta_{\mathrm{x}}$ are absolutely continuous and Lebesgue integrable over the whole domain such that for all $\mathrm{x}\in \mathcal{S}$ we have
     \begin{eqnarray}
    \label{eq:positive}
   0 < \int_{\mathbb{R}^{|\mathrm{x}|}} \zeta_{\mathrm{x}}\, \dd \lambda^{|\mathrm{x}|}  < \infty.
   \end{eqnarray}
   \item[Consistency of Induced Probability Measure:] the membership function induced probability measures $\mathbb{P}_{\zeta_{\mathrm{x}}}$, defined on any $A \in \mathcal{B}(\mathbb{R}^{|\mathrm{x}|})$, as
\begin{IEEEeqnarray}{rCl}
\mathbb{P}_{\zeta_{\mathrm{x}}}(A) & := &  \frac{1}{ \int_{\mathbb{R}^{|\mathrm{x}|}} \zeta_{\mathrm{x}}\, \dd \lambda^{|\mathrm{x}|}} \int_{A} \zeta_{\mathrm{x}}\, \dd\lambda^{|\mathrm{x}|}
\end{IEEEeqnarray}  
are consistent in the sense that for all $\mathrm{x},\;\mathrm{a} \in \mathcal{S}$:
\begin{IEEEeqnarray}{rCl}
\label{eq_738083.390026} \mathbb{P}_{\zeta_{\mathrm{x} \wedge \mathrm{a}}}(A \times \mathbb{R}^{|\mathrm{a}|}) & = & \mathbb{P}_{\zeta_{\mathrm{x}}}(A). 
\end{IEEEeqnarray}   
\end{description}
For convenience, let us denote the collection of membership functions satisfying aforementioned assumptions by 
\begin{IEEEeqnarray}{rCl}
\Theta & := & \{ \zeta_{\mathrm{x}}:\mathbb{R}^{|\mathrm{x}|} \rightarrow [0,1] \; | \; (\ref{eq:supp}),  (\ref{eq:positive}), (\ref{eq_738083.390026}),\; \mathrm{x} \in \mathcal{S}\}.
\end{IEEEeqnarray}
\subsection{A Measure Space}
\begin{result}[A Probability Measure on $\mathbb{F}(\mathcal{X})$]\label{result_probability_space}
$(\mathbb{F}(\mathcal{X}),\sigma(\mathcal{T}),\mathbf{p})$ is a measure space and the probability measure $\mathbf{p}$, that was guaranteed by Kolmogorov extension theorem, is defined as       
 \begin{IEEEeqnarray}{rCl}
\label{eq_738082.742718}\mathbf{p}(\mathcal{T}_{\mathrm{x}}(A) ) & := &  \mathbb{P}_{\zeta_{\mathrm{x}}}(A)
\end{IEEEeqnarray} 
where $\zeta_{\mathrm{x}} \in \Theta$, $\mathrm{x} \in \mathcal{S}$, $A\in \mathcal{B}(\mathbb{R}^{|\mathrm{x}|})$, and $\mathcal{T}_{\mathrm{x}}(A) \in \mathcal{T}$.
\end{result}
\begin{proof}
Given a sequence of samples $(x^i)_{i=1}^{\mathbb{N}}$, define $\mathrm{S}(N) := (x^1,\cdots,x^N)$ i.e. $\mathrm{S}(N+1) = \mathrm{S}(N) \wedge (x^{N+1})$, $N\in \mathbb{N}$. For each $N \in \mathbb{N}$, let $\mathbb{P}_{\zeta_{\mathrm{S}(N)}}$ be a probability measure induced by a membership function $\zeta_{\mathrm{S}(N)} \in \Theta$. As per assumption~(\ref{eq_738083.390026}), the measures, $(\mathbb{P}_{\zeta_{\mathrm{S}(N)}})_{N =1}^{\mathbb{N}}$, are consistent in the sense that $\mathbb{P}_{\zeta_{\mathrm{S}(N+1)}}(A \times \mathbb{R}) = \mathbb{P}_{\zeta_{\mathrm{S}(N)}}(A)$, for any $A \in \mathcal{B}(\mathbb{R}^{N})$ and $N \in \mathbb{N}$. Then Kolmogorov extension theorem guarantees the existence of a probability measure $\mathbf{p}$ on $\mathbb{R}^{\mathbb{N}}$ satisfying $\mathbf{p}(A \times \mathbb{R}^{\mathbb{N}}) = \mathbb{P}_{\zeta_{\mathrm{S}(N)}}(A)$, for any $A \in \mathcal{B}(\mathbb{R}^{N})$. It can be observed that $\mathcal{T} $ forms an algebra of subsets of $\mathbb{F}(\mathcal{X})$. To see this, consider $\mathrm{x} \in \mathcal{S}$, $A \in  \mathcal{B}(\mathbb{R}^{|\mathrm{x}|})$, $\mathrm{a} \in \mathcal{S}$, and $B \in \mathcal{B}(\mathbb{R}^{|\mathrm{a}|})$. Now, we have 
\begin{IEEEeqnarray}{rCl}
\mathbb{F}(\mathcal{X}) & = & \mathcal{T}_{\mathrm{x}}(\mathbb{R}^{|\mathrm{x}|}) \in \mathcal{T}\\
(\mathcal{T}_{\mathrm{x}}(A))^{\complement} & = & \mathcal{T}_{\mathrm{x}}(\mathbb{R}^{|\mathrm{x}|} \setminus A) \in \mathcal{T}\\
\mathcal{T}_{\mathrm{x}}(A) \cap \mathcal{T}_{\mathrm{a}}(B)  & = & \mathcal{T}_{\mathrm{x} \wedge \mathrm{a}}(A \times B) \in \mathcal{T}.
\end{IEEEeqnarray} 
Thus, $\mathcal{T}$ is an algebra of subsets of $\mathbb{F}(\mathcal{X})$. Let $\tilde{\mathbf{p}}:\mathcal{T} \rightarrow [0,1]$ be a function defined as
\begin{IEEEeqnarray}{rCl}   
\label{eq_738082.679662}\tilde{\mathbf{p}}(\mathcal{T}_{\mathrm{x}}(A) )  & := & \mathbb{P}_{\zeta_{\mathrm{x}}}(A).
\end{IEEEeqnarray} 
As $\zeta_{\mathrm{x}} \in \Theta$, (\ref{eq_738083.390026}) holds, and therefore (\ref{eq_738082.679662}) uniquely defines $\tilde{\mathbf{p}}$ over $\mathcal{T}$ without depending on the special representation of cylinder set $\mathcal{T}_{\mathrm{x}}(A)$. It follows from (\ref{eq_738082.679662}) that $\tilde{\mathbf{p}}$ is a $\sigma-$finite \emph{pre-measure} (i.e. $\sigma-$additive) on algebra $\mathcal{T}$ of cylinder sets. Thus, according to \emph{Carath\'eodory's extension theorem}, $\tilde{\mathbf{p}}$ can be extended in a unique way to a measure $\mathbf{p}: \sigma(\mathcal{T}) \rightarrow \mathbb{R}_{\geq 0}$ on the $\sigma-$algebra generated by $\mathcal{T}$. Hence, $(\mathbb{F}(\mathcal{X}),\sigma(\mathcal{T}),\mathbf{p})$ is measure space and a probabilistic measure $\mathbf{p}$, for a set $\mathcal{T}_{\mathrm{x}}(A) \in \mathcal{T}$, is defined as in (\ref{eq_738082.742718}).
\qed
\end{proof}
\begin{result}[Expectations Over $\mathbb{F}(\mathcal{X})$] \label{result_weighted_average}
For a given $\mathcal{B}(\mathbb{R}^{|\mathrm{x}|})-\mathcal{B}(\mathbb{R})$ measurable mapping $g:\mathbb{R}^{|\mathrm{x}|} \rightarrow \mathbb{R}$, expectation of $(g\circ f)(\mathrm{x})$ over $f \in \mathbb{F}(\mathcal{X})$ w.r.t. probability measure $\mathbf{p}$ is given as 
\begin{IEEEeqnarray}{rCl} 
\label{eq_738116.460670} \mathbb{E}_{\mathbf{p}}[(g\circ \cdot)(\mathrm{x})]  & = & \left< g \right >_{\zeta_{\mathrm{x}}}.
\end{IEEEeqnarray} 
\end{result}
\begin{proof}
Given $\mathrm{x} \in \mathcal{S}$, define a projection from $\mathbb{F}(\mathcal{X})$ to $\mathbb{R}^{|\mathrm{x}|}$ as 
\begin{IEEEeqnarray}{rCl}
\Pi_{\mathrm{x}}(f) & := &  f(\mathrm{x})
\end{IEEEeqnarray}
where $f \in \mathbb{F}(\mathcal{X})$. For any $A\in \mathcal{B}(\mathbb{R}^{|\mathrm{x}|})$,
\begin{IEEEeqnarray}{rCl}
\label{eq_738078.807793}\Pi_{\mathrm{x}}^{-1}(A) & = & \mathcal{T}_{\mathrm{x}}(A).
\end{IEEEeqnarray} 
It follows from (\ref{eq_738082.742718}) and (\ref{eq_738078.807793}) that
\begin{IEEEeqnarray}{rCl}
\label{eq_738078.808078}\mathbb{P}_{\zeta_{\mathrm{x}}} & = & \mathbf{p} \circ \Pi_{\mathrm{x}}^{-1}. 
\end{IEEEeqnarray} 
For a $\mathcal{B}(\mathbb{R}^{|\mathrm{x}|})-\mathcal{B}(\mathbb{R})$ measurable mapping $g:\mathbb{R}^{|\mathrm{x}|} \rightarrow \mathbb{R}$, the average value of $g(f(\mathrm{x}))$ over all real valued functions $f\in \mathbb{F}(\mathcal{X})$ can be calculated via taking expectation of $g(\Pi_{\mathrm{x}}(f) )$ w.r.t. probabilistic measure $\mathbf{p}$. That is,
\begin{IEEEeqnarray}{rCl}
\mathbb{E}_{\mathbf{p}}\left[g(f(\mathrm{x}))\right] & = & \mathrm{E}_{\mathbf{p}}\left[g(\Pi_{\mathrm{x}}(f) )\right] \\
& = & \int_{\mathbb{F}(\mathcal{X})} g \circ \Pi_{\mathrm{x}} \: \dd \mathbf{p}\\
& = & \int_{\mathbb{R}^{|\mathrm{x}|}} g \: \dd \mathbb{P}_{\zeta_{\mathrm{x}}}  \\
\label{738080.706409} & = &  \left< g \right >_{\zeta_{\mathrm{x}}}.
\end{IEEEeqnarray} 
\qed 
\end{proof}
\subsection{Student-t Membership-Mapping}
\begin{definition}[Student-t Membership-Mapping]\label{def_student_t_set_membership_mapping}
A Student-t membership-mapping, $\mathcal{F} \in \mathbb{F}(\mathcal{X})$, is a mapping with input space $\mathcal{X} = \mathbb{R}^n$ and a membership function $\zeta_{\mathrm{x}} \in \Theta$ that is Student-t like:
\begin{IEEEeqnarray}{rCl}
\label{eq_student_t_membership} 
\label{eq_738098.751419}\zeta_{\mathrm{x}}(\mathrm{y}) & = & \left(1 + 1/(\nu - 2) \left( \mathrm{y} - \mathrm{m}_{\mathrm{y}} \right)^T K^{-1}_{\mathrm{x}\mathrm{x}} \left( \mathrm{y}- \mathrm{m}_{\mathrm{y}}\right) \right)^{-\frac{\nu+|\mathrm{x}|}{2}}
\end{IEEEeqnarray} 
where $\mathrm{x} \in \mathcal{S}$, $\mathrm{y} \in \mathbb{R}^{|\mathrm{x}|}$, $\nu \in \mathbb{R}_{+}\setminus [0,2]$ is the degrees of freedom, $\mathrm{m}_{\mathrm{y}} \in \mathbb{R}^{|\mathrm{x}|}$ is the mean vector, and $K_{\mathrm{x}\mathrm{x}} \in \mathbb{R}^{|\mathrm{x}| \times |\mathrm{x}|}$ is the covariance matrix with its $(i,j)-$th element given as 
\begin{IEEEeqnarray}{rCl}
\label{738026.844153}  (K_{\mathrm{x}\mathrm{x}})_{i,j} & = & kr(x^i,x^j) 
 \end{IEEEeqnarray}  
where $kr: \mathbb{R}^n \times \mathbb{R}^n \rightarrow \mathbb{R}$ is a positive definite kernel function defined as 
\begin{IEEEeqnarray}{rCl}
\label{eq_membership1003_3} kr(x^{i},x^{j}) & = &  \sigma^2 \exp \left(-0.5\sum_{k = 1}^{n} w_{k} \left |  x^{i}_k - x^{j}_k \right |^2\right)
 \end{IEEEeqnarray}  
where $x_k^i$ is the $k-$th element of $x^i$, $\sigma^2$ is the variance parameter, and $w = (w_1,\cdots,w_n)$ with $w_{k} \geq 0$.  
\end{definition} 
\begin{result}\label{result_student_t_membership_is_consistent}
Membership function as defined in (\ref{eq_738098.751419}) satisfies the consistency condition~(\ref{eq_738083.390026})
\end{result}
\begin{proof}
It follows from (\ref{eq_738098.751419}) that
\begin{IEEEeqnarray}{rCl} 
 \int_{\mathbb{R}^{|\mathrm{x}|}} \zeta_{\mathrm{x}}(\mathrm{y}) \: \dd \lambda^{|\mathrm{x}|}(\mathrm{y}) & = & \frac{\Gamma(\nu/2)}{\Gamma((\nu + |\mathrm{x}|)/2)} (\text{pi})^{|\mathrm{x}|/2} (\nu)^{|\mathrm{x}|/2}\left(\frac{\nu-2}{\nu}\right)^{1/2}|K_{\mathrm{x} \mathrm{x}}|^{1/2},   \IEEEeqnarraynumspace    \\
 \frac{\zeta_{\mathrm{x}}(\mathrm{y})}{\int_{\mathbb{R}^{|\mathrm{x}|}} \zeta_{\mathrm{x}}(\mathrm{y}) \: \dd \lambda^{|\mathrm{x}|}(\mathrm{y}) } & = & p_{\mathrm{\mathbf{y}}} \left(\mathrm{y};\mathrm{m}_{\mathrm{y}},K_{\mathrm{x}\mathrm{x}},\nu\right),
 \end{IEEEeqnarray}   
where $p_{\mathrm{\mathbf{y}}} \left(\mathrm{y};\mathrm{m}_{\mathrm{y}},K_{\mathrm{x}\mathrm{x}},\nu\right)$ is the density function of multivariate $t-$distribution with mean $\mathrm{m}_{\mathrm{y}}$, covariance $K_{\mathrm{x}\mathrm{x}}$ (and scale matrix as equal to $((\nu-2)/\nu)K_{\mathrm{x}\mathrm{x}}$), and degrees of freedom $\nu$. Further, we have
\begin{IEEEeqnarray*}{rCl} 
\frac{\zeta_{\mathrm{x} \wedge \mathrm{a}}((\mathrm{y},\mathrm{u}))}{\int_{\mathbb{R}^{|\mathrm{x}|+|\mathrm{a}|}} \zeta_{\mathrm{x} \wedge \mathrm{a}}((\mathrm{y},\mathrm{u})) \: \dd \lambda^{|\mathrm{x}|+|\mathrm{a}|}((\mathrm{y},\mathrm{u})) } & = & p_{(\mathrm{\mathbf{y}}, \mathrm{\mathbf{u}})} \left((\mathrm{y},\mathrm{u});(\mathrm{m}_{\mathrm{y}},\mathrm{m}_{\mathrm{u}}), \left[\begin{IEEEeqnarraybox*}[][c]{,c/c,}  K_{\mathrm{x}\mathrm{x}} & K_{\mathrm{x}\mathrm{a}} \\ 
K_{\mathrm{a}\mathrm{x}} &K_{\mathrm{a}\mathrm{a}}
 \end{IEEEeqnarraybox*} \right],\nu\right). \IEEEeqnarraynumspace 
 \end{IEEEeqnarray*}      
As the marginal distributions of multivariate $t-$distribution are also $t-$distributions~\cite{Nadarajah2005} i.e.
\begin{IEEEeqnarray}{rCl} 
\int_{\mathbb{R}^{|\mathrm{a}|}} p_{(\mathrm{\mathbf{y}},\mathrm{\mathbf{u}})} \left((\mathrm{y},\mathrm{u});(\mathrm{m}_{\mathrm{y}},\mathrm{m}_{\mathrm{u}}), \left[\begin{IEEEeqnarraybox*}[][c]{,c/c,}  K_{\mathrm{x}\mathrm{x}} & K_{\mathrm{x}\mathrm{a}} \\ 
K_{\mathrm{a}\mathrm{x}} &K_{\mathrm{a}\mathrm{a}}
 \end{IEEEeqnarraybox*} \right],\nu\right) \: \dd \lambda^{|\mathrm{a}|}(\mathrm{u}) & = & p_{\mathrm{\mathbf{y}}} \left(\mathrm{y};\mathrm{m}_{\mathrm{y}},K_{\mathrm{x}\mathrm{x}},\nu\right), \IEEEeqnarraynumspace 
 \end{IEEEeqnarray}    
we have
\begin{IEEEeqnarray}{rCl} 
 \frac{\int_{\mathbb{R}^{|\mathrm{a}|}} \zeta_{\mathrm{x} \wedge \mathrm{a}}((\mathrm{y},\mathrm{u})) \: \dd \lambda^{|\mathrm{a}|}(\mathrm{u}) }{\int_{\mathbb{R}^{|\mathrm{x}|+|\mathrm{a}|}} \zeta_{\mathrm{x} \wedge \mathrm{a}}((\mathrm{y},\mathrm{u})) \: \dd \lambda^{|\mathrm{x}|+|\mathrm{a}|}((\mathrm{y},\mathrm{u})) } & = & \frac{\zeta_{\mathrm{x}}(\mathrm{y})}{\int_{\mathbb{R}^{|\mathrm{x}|}} \zeta_{\mathrm{x}}(\mathrm{y}) \: \dd \lambda^{|\mathrm{x}|}(\mathrm{y}) }.  
 \end{IEEEeqnarray} 
For any $A \in \mathcal{B}(\mathbb{R}^{|\mathrm{x}|})$,
\begin{IEEEeqnarray}{rCl} 
 \frac{\int_{A \times \mathbb{R}^{|\mathrm{a}|}} \zeta_{\mathrm{x} \wedge \mathrm{a}}((\mathrm{y},\mathrm{u})) \: \dd \lambda^{|\mathrm{x}|+|\mathrm{a}|}((\mathrm{y},\mathrm{u})) }{\int_{\mathbb{R}^{|\mathrm{x}|+|\mathrm{a}|}} \zeta_{\mathrm{x} \wedge \mathrm{a}}((\mathrm{y},\mathrm{u})) \: \dd \lambda^{|\mathrm{x}|+|\mathrm{a}|}((\mathrm{y},\mathrm{u})) } & = & \frac{\int_{A}\zeta_{\mathrm{x}}(\mathrm{y}) \: \dd \lambda^{|\mathrm{x}|}(\mathrm{y})}{\int_{\mathbb{R}^{|\mathrm{x}|}} \zeta_{\mathrm{x}}(\mathrm{y}) \: \dd \lambda^{|\mathrm{x}|}(\mathrm{y}) }. 
 \end{IEEEeqnarray}  
Thus, (\ref{eq_738083.390026}) is satisfied.   
\qed
\end{proof}
\subsection{Interpolation by Student-t Membership-Mapping}
Let $\mathcal{F} \in \mathbb{F}(\mathbb{R}^n)$ be a zero-mean Student-t membership-mapping. Let $\mathrm{x} = \{x^i \in \mathbb{R}^n \; | \; i \in \{1,\cdots,N\}\}$ be a given set of input points. The corresponding mapping outputs, represented by the vector $\mathrm{f}  :=  (\mathcal{F}(x^1), \cdots, \mathcal{F}(x^N))$, follow
\begin{IEEEeqnarray}{rCl}
\zeta_{\mathrm{x}}(\mathrm{f}) & = & \left(1 + (1/(\nu - 2))  \mathrm{f}^T K^{-1}_{\mathrm{x}\mathrm{x}}  \mathrm{f} \right)^{-\frac{\nu+N}{2}}.
 \end{IEEEeqnarray} 
Let $\mathrm{a} = \{a^m\;|\; a^m \in \mathbb{R}^n,\; m \in \{1,\cdots,M \}  \}$ be the set of auxiliary inducing points. The mapping outputs corresponding to auxiliary inducing inputs, represented by the vector $\mathrm{u}  :=  (\mathcal{F}(a^1) , \cdots , \mathcal{F}(a^M))$, follow     
\begin{IEEEeqnarray}{rCl}
\zeta_{\mathrm{a}}(\mathrm{u}) & = & \left(1 + (1/(\nu - 2))  \mathrm{u}^T K^{-1}_{\mathrm{a}\mathrm{a}}  \mathrm{u} \right)^{-\frac{\nu+M}{2}}
 \end{IEEEeqnarray} 
where $K_{\mathrm{a}\mathrm{a}} \in \mathbb{R}^{M \times M}$ is positive definite matrix with its $(i,j)-$th element given as
\begin{IEEEeqnarray}{rCl}
\label{eq_membership1004_2} \left( K_{\mathrm{a}\mathrm{a}} \right)_{i,j} & = & kr(a^{i},a^{j}) 
 \end{IEEEeqnarray} 
where $kr: \mathbb{R}^n \times \mathbb{R}^n \rightarrow \mathbb{R}$ is a positive definite kernel function defined as in (\ref{eq_membership1003_3}). Similarly, the combined mapping outputs $(\mathrm{f},\mathrm{u})$ follow
\begin{IEEEeqnarray}{rCl}
\zeta_{\mathrm{x} \wedge \mathrm{a}}((\mathrm{f},\mathrm{u})) & = & 
 \left(1+ \frac{1}{\nu - 2}\left( \left[\begin{IEEEeqnarraybox*}[][c]{,c,}\mathrm{f} \\  \mathrm{u}\end{IEEEeqnarraybox*} \right]  \right)^T   \left[\begin{IEEEeqnarraybox*}[][c]{,c/c,}  K_{\mathrm{x}\mathrm{x}} & K_{\mathrm{x}\mathrm{a}} \\ 
K_{\mathrm{a}\mathrm{x}} &K_{\mathrm{a}\mathrm{a}}
 \end{IEEEeqnarraybox*} \right]^{-1}  \left[\begin{IEEEeqnarraybox*}[][c]{,c,}\mathrm{f} \\  \mathrm{u} \end{IEEEeqnarraybox*} \right]   \right)^{-\frac{\nu+N+M}{2}}.
 \end{IEEEeqnarray}   
It can be verified using a standard result regarding the inverse of a partitioned symmetric matrix  that
 \begin{IEEEeqnarray}{rCl}
\nonumber \lefteqn{\frac{\zeta_{\mathrm{x} \wedge \mathrm{a}}((\mathrm{f},\mathrm{u}))}{ \lvert \zeta_{\mathrm{a}}(\mathrm{u})  \rvert ^ {(\nu+N+M)/(\nu+M)}}  }  \\
  \label{eq_property1_1_student_t}  & = & \left( 1 + \frac{(\mathrm{f} -  \bar{m}_{\mathrm{f}})^T   \left( \frac{\nu + (\mathrm{u})^T (K_{\mathrm{a}\mathrm{a}})^{-1} \mathrm{u}  - 2}{\nu + M - 2} \bar{K}_{\mathrm{x}\mathrm{x}} \right)^{-1}(\mathrm{f} -  \bar{m}_{\mathrm{f}})}{\nu + M - 2}  \right)^{-\frac{\nu+M+N}{2}}, \\
\label{eq_738124.571185} \bar{m}_{\mathrm{f}} & = &  K_{\mathrm{x}\mathrm{a}} (K_{\mathrm{a}\mathrm{a}})^{-1}  \mathrm{u} \\     
\bar{K}_{\mathrm{x}\mathrm{x}} & = & K_{\mathrm{x}\mathrm{x}} - K_{\mathrm{x}\mathrm{a}} (K_{\mathrm{a}\mathrm{a}})^{-1} K_{\mathrm{x}\mathrm{a}}^T.
    \end{IEEEeqnarray} 
The expression on the right hand side of equality~(\ref{eq_property1_1_student_t}) define a Student-t membership function with the mean $ \bar{m}_{\mathrm{f}}$. It is observed from (\ref{eq_738124.571185}) that $ \bar{m}_{\mathrm{f}}$ is an interpolation on the elements of $\mathrm{u}$ based on the closeness of points in $\mathrm{x}$ with that of $\mathrm{a}$. Hence, $\mathrm{f}$, based upon the interpolation on elements of $\mathrm{u}$, could be represented by means of a membership function, $\mu_{\mathrm{f};\mathrm{u}}:\mathbb{R}^N \rightarrow [0,1]$, defined as r.h.s. of (\ref{eq_property1_1_student_t}):
 \begin{IEEEeqnarray}{rCl}
\mu_{\mathrm{f};\mathrm{u}}(\tilde{\mathrm{f}})  & := & \left( 1 + \frac{(\tilde{\mathrm{f}} -  \bar{m}_{\mathrm{f}})^T    \left( \frac{\nu + (\mathrm{u})^T (K_{\mathrm{a}\mathrm{a}})^{-1} \mathrm{u}  - 2}{\nu + M - 2} \bar{K}_{\mathrm{x}\mathrm{x}} \right)^{-1}(\tilde{\mathrm{f}} -  \bar{m}_{\mathrm{f}})}{\nu + M - 2}  \right)^{-\frac{\nu+M+N}{2}}. \IEEEeqnarraynumspace 
\end{IEEEeqnarray}    
Here, the pair $(\mathbb{R}^N,\mu_{\mathrm{f};\mathrm{u}})$ constitutes a fuzzy set and $\mu_{\mathrm{f};\mathrm{u}}(\tilde{\mathrm{f}})$ is interpreted as the degree to which $\tilde{\mathrm{f}}$ matches an attribute induced by $\mathrm{f}$ for a given $\mathrm{u}$.       
\subsection{Variational Learning of Membership-Mappings}
 Given a dataset $\{(x^i,y^i)\;|\; x^i \in \mathbb{R}^n,\;y^i \in \mathbb{R}^p,\; i \in \{1,\cdots,N \} \}$, it is assumed that there exist zero-mean Student-t membership-mappings $\mathcal{F}_1, \cdots, \mathcal{F}_p \in \mathbb{F}(\mathbb{R}^n)$ such that
\begin{IEEEeqnarray}{rCl}
\label{eq_738118.641846} y^i &\approx & \left[\begin{IEEEeqnarraybox*}[][c]{,c/c/c,} \mathcal{F}_1(x^i)  & \cdots & \mathcal{F}_p(x^i) \end{IEEEeqnarraybox*} \right]^T.
\end{IEEEeqnarray} 
Under modeling scenario~(\ref{eq_738118.641846}), a variational learning solution can be derived via following an analytical approach as in~\cite{8888203,9216097,KUMAR20211}. Representing the variables associated to a membership-mapping model by means of membership functions, the mathematical expressions for membership functions are analytically derived using variational optimization such that the degree-of-belongingness of given data to the considered model is maximized. The analytical approach leads to the development of Algorithm~\ref{algorithm_basic_learning} for learning. With reference to Algorithm~\ref{algorithm_basic_learning},   
\begin{itemize}
\item $\mathrm{y}_j $, for $ j \in \{1,2,\cdots,p\}$, is defined as 
\begin{IEEEeqnarray}{rCl}
\label{eq_y_j_vec2000} \mathrm{y}_j & := & \left[\begin{IEEEeqnarraybox*}[][c]{,c/c/c,}y_j^1 & \cdots &y_j^N\end{IEEEeqnarraybox*} \right]^T \in \mathbb{R}^N
 \end{IEEEeqnarray}  
 where $y_j^i$ denotes the $j-$th element of $y^i$.
\item $\xi$ is given as
\begin{IEEEeqnarray}{rCl}
\label{eq_xi2000_l} \xi & = &  N \sigma^2.  
\end{IEEEeqnarray}  
\item $\Psi \in \mathbb{R}^{N \times M}$ is a matrix with its $(i,m)-$th element given as
\begin{IEEEeqnarray}{rCl}
\label{eq_psi2000_l}  \Psi_{i,m} & = & \frac{\displaystyle \sigma^2  }{ \prod_{k=1}^{n}\left(\sqrt{1 + w_{k} \sigma_x^2}\right)} \exp\left(-\frac{1}{2} \sum_{k=1}^{n} \frac{w_{k} |a^{m}_k - x^i_k |^2}{1 + w_{k} \sigma_x^2} \right) 
\end{IEEEeqnarray}  
where $a^{m}_k$ and $x^i_k$ denotes the $k-$th element of $a^{m}$ and $x^i$ respectively. 
\item $\Phi \in \mathbb{R}^{M \times M}$ is a matrix with its $(m,m^{\prime})-$th element given as
\begin{IEEEeqnarray}{rCl}
\nonumber \Phi_{m,m^{\prime}} & = & \frac{\displaystyle \sigma^4 }{ \prod_{k=1}^{n}\left( \sqrt{1 + 2w_{k} \sigma_x^2 } \right)}   \sum_{i=1}^N \exp\left(\displaystyle -\frac{1}{4} \sum_{k=1}^{n} w_{k} (a^{m}_k - a^{m^{\prime}}_k)^2  \right. \\
\label{eq_phi2000_l}  && \left. - \sum_{k=1}^{n} \frac{w_{k}  |0.5(a^{m}_j + a^{m^{\prime}}_k) - x^i_k |^2}{1 + 2w_{k}\sigma_x^2} \right).   \IEEEeqnarraynumspace  
\end{IEEEeqnarray}  
\item The quantities $(\hat{a}_{\tau},\hat{b}_{\tau},\hat{a}_{z},\hat{b}_{z},\hat{a}_{r},\hat{b}_{r},\hat{a}_{s},\hat{b}_{s})$ follow
\begin{IEEEeqnarray}{rCl}
\label{eq_FTNDL_1_update_lambda_l} \hat{a}_{\tau} & = & a_{\tau} +  0.5Np \\
 \label{eq_FTNDL_2_update_lambda_l} \hat{b}_{\tau}(O) & = & b_{\tau}   +  \frac{\hat{a}_{z}}{2\hat{b}_{z}}O
\end{IEEEeqnarray}
  \begin{IEEEeqnarray}{rCl}
\label{eq_FTNDL_1_update_z} \hat{a}_{z} & = & 1 + 0.5Np  + \hat{a}_{r}/\hat{b}_{r}  \\
 \label{eq_FTNDL_2_update_z} \hat{b}_{z}(O) & = & \frac{\hat{a}_{r}}{\hat{b}_{r}}\frac{\hat{a}_{s}}{\hat{b}_{s}}    + \frac{\hat{a}_{\tau}}{2\hat{b}_{\tau}} O
\end{IEEEeqnarray}
\begin{IEEEeqnarray}{rCl}
\label{eq_FTNDL_1_update_r} \hat{a}_{r} & = & a_{r} \\
 \label{eq_FTNDL_2_update_r}  \hat{b}_{r} & = & b_{r} +  (\hat{a}_{s}/\hat{b}_{s})  (\hat{a}_{z}/\hat{b}_{z})  -  \psi\left(\hat{a}_{s}\right) + \log\left(\hat{b}_{s}\right)    - 1 -  \psi\left(\hat{a}_{z}\right) + \log\left(\hat{b}_{z}\right) 
\end{IEEEeqnarray}
 \begin{IEEEeqnarray}{rCl}
\label{eq_FTNDL_1_update_s}  \hat{a}_{s} & = & a_{s} +  (\hat{a}_{r}/\hat{b}_{r}) \\
 \label{eq_FTNDL_2_update_s} \hat{b}_{s} & = & b_{s} +  (\hat{a}_{r}/\hat{b}_{r})    (\hat{a}_{z}/\hat{b}_{z})
\end{IEEEeqnarray}
\end{itemize}
\begin{algorithm}
\caption{Variational learning of the membership-mappings}
\label{algorithm_basic_learning}
\footnotesize{
\begin{algorithmic}[1]
\REQUIRE  Dataset $\left\{ (x^i,y^i) \; | \; x^i \in \mathbb{R}^n,\; y^i \in \mathbb{R}^p,\; i \in \{1,\cdots,N \} \right \}$; number of auxiliary points $M \in \{1,2,\cdots,N\}$; the degrees of freedom associated to the Student-t membership-mapping $\nu \in \mathbb{R}_{+} \setminus [0,2]$.    
\STATE Choose free parameters as $\sigma^2 = 1$ and $\sigma_x^2 = 0.01$. 
\STATE The auxiliary inducing points are suggested to be chosen as the cluster centroids: 
\[
\mathrm{a} = \{ a^{m}\}_{m=1}^M  =  cluster\_centroid(  \{x^i\}_{i=1}^N, M) 
 \]  
where $cluster\_centroid(  \{ x^i \}_{i=1}^N,M)$ represents the k-means clustering on $ \{ x^i \}_{i=1}^N$. 
\STATE Define $w = (w_1,w_2,\cdots,w_n)$ such that $w_{k}$ (for $k\in \{1,2,\cdots,n\}$) is equal to the inverse of squared-distance between two most-distant points in the set: $\{ x^{1}_k, x^{2}_k,\cdots, x^{N}_k \}$.
\STATE Compute $K_{\mathrm{a}\mathrm{a}}$, $\xi$, $\Psi$, and $\Phi$ using (\ref{eq_membership1004_2}), (\ref{eq_xi2000_l}), (\ref{eq_psi2000_l}), and (\ref{eq_phi2000_l}) respectively.
\STATE Choose $a_{\tau} = b_{\tau} = a_{r} = b_{r} = a_{s} = b_{s} = 1$.
\STATE Initialise $\hat{a}_{\tau} = \hat{b}_{\tau} = \hat{a}_{z} = \hat{b}_{z} = \hat{a}_{r} = \hat{b}_{r} = 1$.  
\STATE Initialize $\hat{a}_{s}$ and $\hat{b}_{s}$ using (\ref{eq_FTNDL_1_update_s}) and (\ref{eq_FTNDL_2_update_s}). 
 \REPEAT
\STATE Update $\mathcal{E}(\hat{m}_{\mathrm{u}_j}(\mathrm{y}_j))$ as
\begin{IEEEeqnarray}{rCl}
\label{eq_final_hat_m_u_1000} \mathcal{E}(\hat{m}_{\mathrm{u}_j}(\mathrm{y}_j))  & = &   K_{\mathrm{a}\mathrm{a}} \left(   \Phi  +   \frac{\xi- Tr((K_{\mathrm{a}\mathrm{a}})^{-1}  \Phi)}{\nu+ M - 2}  K_{\mathrm{a}\mathrm{a}}  +  \frac{\hat{b}_{\tau} \hat{b}_{z}}{\hat{a}_{\tau} \hat{a}_{z}}  K_{\mathrm{a}\mathrm{a}} \right)^{-1} (\Psi)^T \mathrm{y}_j. 
\end{IEEEeqnarray}   
\STATE Update $\mathcal{E}(O)$ as
\begin{IEEEeqnarray}{rCl}
 \nonumber \mathcal{E}(O) & = & \sum_{j=1}^p \left( \| \mathrm{y}_j \|^2 - 2 \left(\mathcal{E}(\hat{m}_{\mathrm{u}_j}(\mathrm{y}_j))\right)^T (K_{\mathrm{a}\mathrm{a}})^{-1} (\Psi)^T \mathrm{y}_j     \right. \\
\nonumber && \left. {+}\: \left(\mathcal{E}(\hat{m}_{\mathrm{u}_j}(\mathrm{y}_j)) \right)^T (K_{\mathrm{a}\mathrm{a}})^{-1}  \Phi (K_{\mathrm{a}\mathrm{a}})^{-1}  \mathcal{E}(\hat{m}_{\mathrm{u}_j}(\mathrm{y}_j)) \right. \\
\label{eq_satguru_17}  && \left.  {+}\:  \left(\mathcal{E}(\hat{m}_{\mathrm{u}_j}(\mathrm{y}_j)) \right)^T  \frac{\xi - Tr((K_{\mathrm{a}\mathrm{a}})^{-1}  \Phi)}{\nu+ M - 2} (K_{\mathrm{a}\mathrm{a}})^{-1} \mathcal{E}(\hat{m}_{\mathrm{u}_j}(\mathrm{y}_j))\right). \IEEEeqnarraynumspace
 \end{IEEEeqnarray} 
\STATE Update $\hat{a}_{\tau},\hat{b}_{\tau}(\mathcal{E}(O) ),\hat{a}_{z},\hat{b}_{z}(\mathcal{E}(O) ),\hat{a}_{r},\hat{b}_{r},\hat{a}_{s},\hat{b}_{s}$ using (\ref{eq_FTNDL_1_update_lambda_l}), (\ref{eq_FTNDL_2_update_lambda_l}), (\ref{eq_FTNDL_1_update_z}), (\ref{eq_FTNDL_2_update_z}), (\ref{eq_FTNDL_1_update_r}), (\ref{eq_FTNDL_2_update_r}), (\ref{eq_FTNDL_1_update_s}), (\ref{eq_FTNDL_2_update_s}) respectively.  
\STATE Estimate $\beta$ as
 \begin{IEEEeqnarray}{rCl}
\label{eq_added_2_23_07_2020} \beta & = &  (\hat{a}_{\tau}/\hat{b}_{\tau})  (\hat{a}_{z}/\hat{b}_{z}).
 \end{IEEEeqnarray} 
 \UNTIL{($\beta$ nearly converges)}
\STATE Compute matrix $B$ as
  \begin{IEEEeqnarray}{rCl}
\label{eq_738128.804979} B & =&  \left( \Phi+ \frac{\xi- Tr((K_{\mathrm{a}\mathrm{a}})^{-1}  \Phi)}{\nu+ M - 2}  K_{\mathrm{a}\mathrm{a}} +  \frac{\hat{b}_{\tau}}{\hat{a}_{\tau}}  \frac{\hat{b}_{z}}{\hat{a}_{z}} K_{\mathrm{a}\mathrm{a}}\right)^{-1}   (\Psi)^T.
 \end{IEEEeqnarray} 
Compute matrix $\alpha = \left[\begin{IEEEeqnarraybox*}[][c]{,c/c/c,}  \alpha_1 & \cdots & \alpha_p
 \end{IEEEeqnarraybox*} \right]$ with its $j-$th column defined as 
\begin{IEEEeqnarray}{rCl}
\label{eq_vector_alpha}  \alpha_j & := & \left( \Phi+ \frac{\xi- Tr((K_{\mathrm{a}\mathrm{a}})^{-1}  \Phi)}{\nu+ M - 2}  K_{\mathrm{a}\mathrm{a}} +  \frac{\hat{b}_{\tau}}{\hat{a}_{\tau}}  \frac{\hat{b}_{z}}{\hat{a}_{z}} K_{\mathrm{a}\mathrm{a}}\right)^{-1}   (\Psi)^T \mathrm{y}_j 
  \end{IEEEeqnarray} 
\RETURN  The parameters set $\mathbb{M} = \{\alpha, w,  \mathrm{a} , \sigma^2, \sigma_x^2,B\}$.
\end{algorithmic} }
\end{algorithm}     
\subsection{Prediction by Membership-Mappings}
Given the parameters set $\mathbb{M} = \{\alpha, w,  \mathrm{a} , \sigma^2, \sigma_x^2,B\}$ returned by Algorithm~\ref{algorithm_basic_learning}, the learned membership-mappings could be used to predict output corresponding to any arbitrary input data point $x^* \in \mathbb{R}^n$ as
\begin{IEEEeqnarray}{rCl}
\label{eq_738124.770095}\hat{y}(x^*;\mathbb{M}) & = & \alpha^T(G(x^*;\mathbb{M}))^T.
\end{IEEEeqnarray}
Here, $G \in \mathbb{R}^{1 \times M}$ is a vector-valued function defined as
\begin{IEEEeqnarray}{rCl}
\label{eq_738268.700344} G(x;\mathbb{M}) & := & \left[\begin{IEEEeqnarraybox*}[][c]{,c/c/c,} G_1(x;\mathbb{M}) & \cdots &  G_M(x;\mathbb{M})
 \end{IEEEeqnarraybox*} \right] \IEEEeqnarraynumspace  \\
\label{eq_satguru_9} G_m(x;\mathbb{M})& := & \frac{\displaystyle \sigma^2  }{ \prod_{k=1}^{n}\left(\sqrt{1 + w_{k} \sigma_x^2}\right)} \exp\left(-\frac{1}{2} \sum_{k=1}^{n} \frac{w_{k} |a^{m}_k - x_k |^2}{1 + w_{k} \sigma_x^2} \right), \IEEEeqnarraynumspace  
\end{IEEEeqnarray}   
where $a^{m}_k$ and $x_k$ are the $k-$th elements of $x$ and $a^{m}$ respectively. 
\section{Concluding Remarks}
This paper has introduced the notion of membership-mapping using measure theoretic basis for representing data points through attribute values.

\bibliographystyle{splncs04}
\bibliography{bibliography}

\end{document}